# Diffusion-Weighted Magnetic Resonance Brain Images Generation with Generative Adversarial Networks and Variational Autoencoders: A Comparison Study

**Alejandro Ungría Hirte, BSc**
Institute for Biomedical Engineering
ETH Zurich and University of Zurich

**Moritz Platscher, PhD**
Institute for Biomedical Engineering
ETH Zurich and University of Zurich

**Thomas Joyce, PhD**
Institute for Biomedical Engineering
ETH Zurich and University of Zurich

**Jeremy J. Heit, MD, PhD**
Stanford Stroke Center
Department of Neurology
Stanford University

**Eric Tranvinh, MD**
Stanford Stroke Center
Department of Neurology
Stanford University

**Christian Federau, MD, MSc**
Institute for Biomedical Engineering
ETH Zurich and University of Zurich

## Abstract

We show that high quality, diverse and realistic-looking diffusion-weighted magnetic resonance images can be synthesized using deep generative models. Based on professional neuroradiologists' evaluations and diverse metrics with respect to quality and diversity of the generated synthetic brain images, we present two networks, the Introspective Variational Autoencoder and the Style-Based GAN, that qualify for data augmentation in the medical field, where information is saved in a dispatched and inhomogeneous way and access to it is in many aspects restricted.

Corresponding authors : Moritz Platscher platscher@biomed.ee.ethz.ch, Christian Federau federau@biomed.ee.ethz.ch

# 1 Introduction

Deep learning methods have the potential to relieve the radiologist from handling tedious, time-consuming tasks such as detecting and segmenting pathological lesions [1], but the training of neural networks in the context of medical imaging faces a major challenge: they require a large number of images to train, which is difficult to obtain because access to medical information is in many aspects restricted due to patient privacy and legal matters among a wide variety of obstacles. Further, medical images are saved in a dispatched and inhomogeneous way in the different institutions and gathering them in a larger database is challenging.

In this context, a method to generate medical images from scratch might be of great interest. Generative modeling is a subfield of machine learning that has been impressively proficient in generating new high-quality natural images, for example photographs of faces [2], but also at tasks such as speech synthesis [3]. If generative models could be shown to be able to produce realistic and diverse new medical images, they would have the appealing potential to significantly increase the number of images available for deep neural network training at almost no cost, and could therefore help to improve the accuracy of those networks [4]–[6].

The main idea behind generative models is that they should learn a probability distribution over images which captures the (implicit) probability distribution over the training data. One approach is to learn a lower dimensional representation (a so-called *latent space*) of the training data and enforce a simple probability distribution over this latent space. New data can then be generated by sampling from this lower dimensional representation. Following this approach, Variational Autoencoders [7]–[9] (VAEs) and Generative Adversarial Networks [10, 11] (GANs) are two dominant approaches among the multiple types of generative models that have been proposed and extensively studied in recent years. The two approaches have their own strengths and weaknesses: VAEs are known to generate blurry images [9, 12, 13] while GANs are known to be able to generate sharper images, but might lack sample diversity [14] and might be more difficult to train.

The purpose of this study is to investigate the potential to generate realistic new magnetic resonance diffusion-weighted images (DWIs) with four current state of the art generative neural networks based on the GAN and VAE architectures: a vanilla VAE, a *Deep Feature Consistent VAE* [15]*,* an *Introspective VAE* [16] and a *Style-Based GAN*

[17]. The latter three networks offer modifications to the original structures to reduce the above mentioned drawbacks of VAEs and GANs: In short (see the background theory section below for more details), the *Deep Feature Consistent VAE* [15] introduces the idea of comparing original and reconstructed images in the feature space rather than in the pixel space, enforcing a spatial correlation able to produce synthetic data of higher quality compared to a vanilla VAE; the *Introspective Variational Autoencoder* [16] is a hybrid between a VAE and a GAN and suggests to combine GANs' high quality image generations and VAEs' stable training properties; the *Style-Based GAN* [17] tries to alleviate the low diversity problem and to improve control over the latent variables. In the present study, we evaluate the quality, variability and diversity of the newly generated DWIs of those four networks in a quantitative manner. To this end, we introduce four new metrics to evaluates sample diversity and image quality of the generated images, since existing metrics were developed for natural images. Finally, we assess if the generated images are realistic enough to fool two expert neuroradiologists from an external institution fully blinded to the intentions of the study.

## 2 Background Theory

The term "*generative model*" is being used in many different ways. In this paper, the term assumes that the data under consideration is sampled from an underlying probability distribution. This underlying probability distribution is not known, and the generative model tries to learn, in one way or another, an estimate of that distribution [7], [18]. The various generative models differ in the way this distribution is found. To appreciate the complexity of this task, imagine that every voxel could either be white or black. The number of possibilities to generate MRI volumes of size $128 \times 128 \times 40$ is incomprehensibly large: the number of possible configurations reaches an unimaginable $2^{128 \times 128 \times 40} \approx 10^{10^5}$. Trying to generate artificial samples without dimension reduction would easily exhaust even the most powerful computational resources.

However, not all voxels are independent: their values depend on neighboring voxels, global features, and the local structure to which they belong. In order to capture these local and global correlations, we assume that the features contained in the data can be represented in a lower-dimensional *latent space* $\mathcal{L} \subseteq \mathbb{R}^m$ and can be mapped to the data space, e.g. $\mathbb{R}^{n \times n}$ for 2D $n \times n$ black and white images, where $m \ll n^2$. The task of the generative model is now reduced to finding a suitable approximation of the probability distribution over this latent space, which is a considerably less complex task, and to which several approaches exist.

The generative models used in the work are based on Variational Autoencoders (VAEs) and Generative Adversarial Networks (GANs), which use different mathematical frameworks to find this representation. VAEs use an explicit probability function inside a probabilistic graphical model and maximize a lower bound on the log likelihood of the data, while GANs use an implicit probability function which is found through an adversarial game between two separate networks. We first introduce VAEs and GANs, then the three derived networks used here.

**2.1 Variational Autoencoders (VAEs)** [7, 8] first compress an input sample $x \in \mathbb{R}^n$ into latent space via an encoder $E: x \mapsto z \in \mathcal{L}$, followed by a mapping back to the data space, e.g. the pixels of an image, via a decoder (or generator) network. The encoder is a parametric family of approximations to the conditional posterior of a latent vector given the data $q_\theta(z|x) \approx p(z|x)$. The decoder, on the other hand, is a parametric approximation of the likelihood of the data given a latent vector, $p_\phi(x|z) \approx p(x|z)$. Assuming that the true distribution is contained in the parametric family for $\phi = \phi^*$, we are interested in finding a distribution that best represents the data, or in other words, in finding $\phi^*$ that maximizes the true marginalized posterior of the data

$$p(x) \approx p_{\phi^*}(x) = \int_{\mathcal{L}} dz\, p(z)\, p_{\phi^*}(x|z), \tag{1}$$

Where $p(z)$ is a prior. In many cases this integral is computationally intractable. However, a tractable lower bound of $p_\phi(x)$ can be obtained by realizing that the first term in

$$\log\left(p_\phi(x)\right) = D_{KL}[q_\theta(z|x)||p_\phi(z)] + L \geq L \tag{2}$$

is strictly non-negative, where

$$D_{KL}[q(z)||p(z)] \equiv \int dz\, q(z)\log\left(\frac{q(z)}{p(z)}\right) \tag{3}$$

is the Kullback-Leibler divergence. The lower bound is

$$L = -D_{KL}[q_\theta(z|x)||p(z)] + \mathbb{E}_{z \sim q_\theta(z|x)}[p_\phi(x|z)]. \tag{4}$$

Maximizing $L$ w.r.t. the autoencoder's parameters $\theta, \phi$ will result in an approximation of the true log-likelihood, cf. Eq. (2). In practice, the last term in Eq. (4) is often taken to be a reconstruction error, e.g. a pixel-wise $L_2$ loss [9], and the first term ensures that the approximate posterior $q_\theta(z|x)$ is close to some prior distribution, often chosen to be a symmetric Gaussian with mean 0 and unit covariance, $p(z) = \mathcal{N}(z; 0, \mathbb{I})$.

In practice, the encoder and decoder are represented by two independent neural networks, q and p with parameter $\theta$ and $\phi$ respectively. The encoder will map an input image to latent space according to the approximate posterior, $E(x;\theta) \sim q_\theta(z|x)$, and therefore must represent some *explicit* probability distribution, e.g. a normal distribution with mean $\mu$ and covariance $\Sigma$, $\mathcal{N}(\mu, \Sigma)$. The decoder will then be passed a latent vector that is sampled from this distribution via the reparameterization trick [7], $z = \mu + \varepsilon \odot \Sigma$, with an auxiliary noise variable $\varepsilon \sim \mathcal{N}(0,1)$. This step is necessary in order for backpropagation to work properly, while still injecting some stochasticity into the generative part of the network [7]. Finally, the decoder maps this latent vector back to pixel space, $x' = D(z;\phi)$, as illustrated in Fig. 1a). The two networks are trained together to minimize the loss function:

$$\mathcal{L}_{VAE} = D_{KL}[q_\theta(z|x)||p(z)] + ||x - x'||_2^2. \tag{5}$$

Once training has converged, we can draw new latent vectors from the prior, $z \sim p(z)$, and map them to pixel space using the decoder. The training objective (Equation 5) ensures that the sampling will produce realistic artificial data by keeping the true posterior close to the prior during training. While this approach allows us to map certain features in the data into the latent space, e.g. by using cluster algorithms or conditioning on some input labels, it suffers from blurry image generation due to the stochastic nature of the sampling process, the pixel-wise reconstruction error [9, 16], and the variational training objective (Equation 5) itself [12].

**2.2 Generative Adversarial Networks (GANs)** [10, 11, 18], in contrast to VAEs, do not require an explicit approximation to the probability density. A generator network is fed an input sample from a prior distribution in latent space, $\mathcal{L} \ni z \sim p(z)$, and maps these latent vectors to the data space, $G: \mathcal{L} \to \mathbb{R}^n$ (see also Fig. 1b). An accompanying discriminator network is required during training. The discriminator network is a classifier that is trained to differentiate real data samples $\vec{x}$ from samples generated by the generator, $x' = G(z;\theta_G)$, where $\theta_G$ represent the parameters of the generator. Typically, the discriminator represents the probability of an image being generated by $G$ in which case we have $D: \mathbb{R}^n \to [0,1]$. Both networks compete against each other and are therefore trained jointly. The optimal parameters for both network are found in adversarial min-max game [10],

$$\theta_G, \theta_D = \underset{G}{\text{argmin}}\, \underset{D}{\text{argmax}}\, \mathbb{E}_{x \sim p_{data}(x)}[\log D(x;\theta_D)] + \mathbb{E}_{z \sim p(z)}[\log(1 - D(G(z;\theta_G);\theta_D))], \tag{5}$$

where $\theta_D$ represent the parameters of the discriminator. By training the discriminator to better distinguish real samples from generated samples, and the generator to produce ever more realistic samples, GANs learn to sample from the true probability distribution of the data without knowing its explicit form [18]. Thus, GANs may in principle represent the true data distribution, since they do not require a (variational) lower bound as opposed to VAEs [18]. However, they are prone to mode collapse if the true distribution is multi-modal [14] and known for unstable training dynamics [19, 20]. In spite of all these shortcomings, GANs are generally recognized to produce images that are more realistic than VAEs, and a number of methods have been developed that help stabilize training, increase variability, and better capture all modes present in the data.

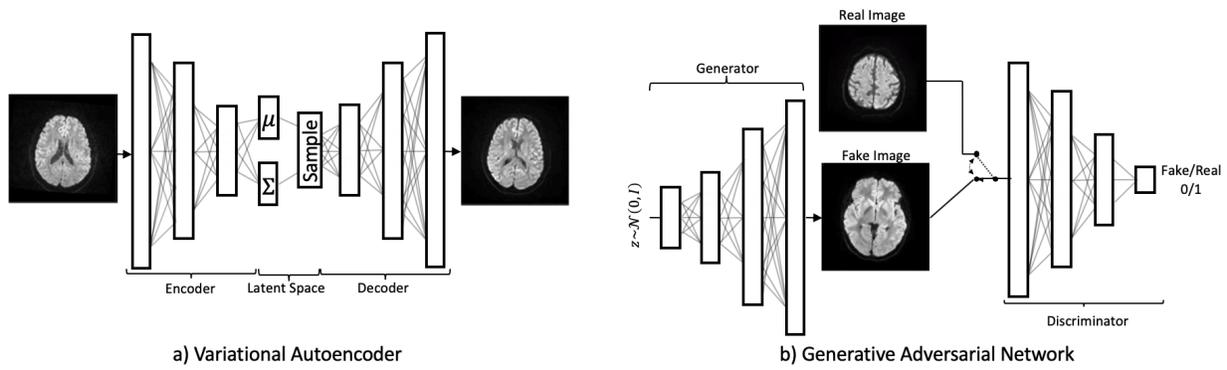

*Figure 1. a) Variational Autoencoder. Input images are encoded to a lower dimensional latent vector $z \sim \mathcal{N}(\mu, \Sigma)$. Minimization of Kullback-Leibler Divergence and MSE will push $z \sim \mathcal{N}(\mu, \Sigma) \approx \mathcal{N}(0, I)$ while still capturing enough information of the pixel space for an accurate reconstruction of the original image. After training, feeding a vector $z \sim \mathcal{N}(0, I)$ to the decoder results in synthetic MRIs. b) Generative Adversarial Network. Discriminator is trained by classifying original and synthetic brains alternatingly. Generator optimized until able to create synthetic MRIs from unit gaussian distributed vectors that fool discriminator.*

### 2.3 Deep Feature Consistent Variational Autoencoder (DFC VAE)

The *DFC VAE* was introduced in Ref. [15] to improve the quality of images generated in VAEs, while retaining their ability to map features into latent space. To this end, the authors replaced the pixel-wise loss with a so-called *perceptual loss*, while keeping the KL-loss in order to ensure that the training image distribution is close to the prior. The reasoning behind replacing the pixel-wise loss was the insight that the same image offset by a few pixels has little visual perceptual difference for humans but can have a very high pixel-wise loss. As a consequence, images generated from VAEs tend to be very blurry when compared to natural images. The perceptual loss implemented in the *DFC VAE* is the mean squared error loss of the hidden feature representations of original and reconstructed images encoded in specific deep layers of a pretrained deep convolutional network $\psi$ (here a VGG16 [21] pretrained on ImageNet [22] was used), using the assumptions that those layers contain more conceptual information of the images compared to the pixel-wise comparison. The *DFC VAE* replaces the second term in Eq. (5) with a term

$$\mathcal{L}_{rec} = \sum_{l=1}^{n} \mathcal{L}_{rec}^{l} \tag{6}$$

where $n$ is the depth, i.e. the number of layers of $\psi$, and

$$\mathcal{L}_{rec}^{l} = \frac{1}{2 H_l W_l C_l} \sum_{c=1}^{C^l} \sum_{w=1}^{W^l} \sum_{h=1}^{H^l} \left(\psi(x)_{c,w,h}^{l} - \psi(x')_{c,w,h}^{l}\right)^2 \tag{7}$$

the perceptual loss at layer $l$, given by the mean square error of the output of $\psi$ for a training image x and its reconstruction $x'$. Moreover, $H^l, W^l$ and $C^l$ are the height, width, and number of filters of the output at this layer, respectively.

## 2.4 Introspective Variational Autoencoder (introVAE)

The *introVAE* [16] adds a GAN-like discriminator to the architecture of the vanilla VAE: the encoder is both a mapping from training data to latent space, and simultaneously acts as discriminator (Fig. 2b). Intuitively, if the generator, out of random sampling from the latent space, is capable of generating images that can be encoded to latent vectors with the same latent space distribution as the encodings of original images, it should have learnt well the underlying data distribution. Thus, when the encoder receives an original image x, it is trained to produce an encoding that follows a standard gaussian as in the original VAE framework by minimizing their KL-divergence. In the case of receiving a synthetic image $x' = G(z)$, it is conversely trained to find a latent representation that maximizes the KL-divergence (or equivalently minimizes the negative KL-divergence). This translates into minimizing, simultaneously for real images $x$ and fake images $x' = G(z)$,

$$\mathcal{L}_E(x,z) = \mathcal{L}_{KL}\big(E(x)\big) + [m - \mathcal{L}_{KL}(E(x'))]^+ + \mathcal{L}\left(x, G\big(E(x)\big)\right) \tag{8}$$

to train the encoder. Here, $\mathcal{L}_{KL}$ is KL-divergence between the distribution of latent variables and the prior, e.g. $p(z) = \mathcal{N}(0,1)$; $\mathcal{L}_{rec}(x, \vec{y})$ is the mean squared error between two images $x$ and $\vec{y}$ and $[\cdot]^+ = \max(0,\cdot)$. Conversely, the generator must enforce the reconstructions to match the input and simultaneously learn to generate synthetic images from the prior distribution of latent vectors:

$$\mathcal{L}_G(z) = \mathcal{L}_{KL}\big(E(G(z))\big) + \mathcal{L}_{rec}\big(x, G(E(x))\big). \tag{9}$$

**2.5 *Style-Based Generative Adversarial Network (styleGAN)***

The *styleGAN* [17] is a variation of GANs with a modified generator input: instead of mapping a latent vector to pixel space directly, the generator takes a constant input, and passes it through several residual blocks [23], which receive latent vector inputs through so-called *adaptive instance normalization* (AdaIN) layers [17],

$$\text{AdaIN}(x_i, y) = y_{s,i} \frac{x_i - \mu(x_i)}{\sigma(x_i)} + y_{b,i}, \tag{10}$$

where $x_i$ is the input, $\mu(x_i)$ and $\sigma(x_i)$ are mean and standard deviation calculated from $x_i$, and $y = (y_{s,i}, y_{b,i})$ are the learnt style inputs. These styles are obtained in the following way (see also Fig. 2c): First, a latent vector is sampled from a simple prior, $z \sim p(z)$, and passed through a network of fully connected layers to allow for more general distributions. The resulting style vector $w$ is then passed to the AdaIn layers through linear transformations that yield tuples $(y_{s,i}, y_{b,i})$. Additionally, noise is integrated at each layer to introduce style variation at different levels of detail. Moreover, the *styleGAN* is trained progressively, first on training images downsampled to $4 \times 4$ pixels, then subsequently doubling their size until the original resolution is reached. This has been shown to speed-up training and improve convergence [17]. Finally, instead of the training objective in Eq. (5), the *styleGAN* is trained to minimize the so-called Wasserstein distance [24],

$$\mathcal{L}_D = \mathbb{E}_{x \sim p_{data}(x)}[D(x; \theta_D)] - \mathbb{E}_{z \sim p_z(z)}[D(G(z; \theta_G); \theta_D)], \tag{11}$$

$$\mathcal{L}_G = \mathbb{E}_{z \sim p_z(z)}[D(G(z; \theta_G); \theta_D)]. \tag{12}$$

Ref. [17] showed that, once the discriminator has been trained to optimality, the generator objective in Eq. (5) is equivalent to minimizing the Jensen-Shannon divergence between the model and the data distribution. However, it has been argued that the Wasserstein distance [which amounts to the above training objectives (11,12)] is a more suitable measure of similarity, should prevent mode collapse and stabilize the training [24].

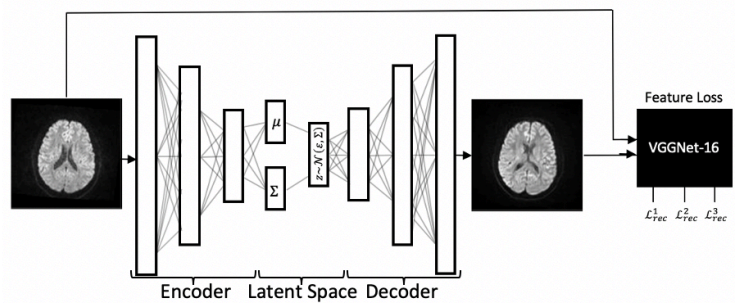
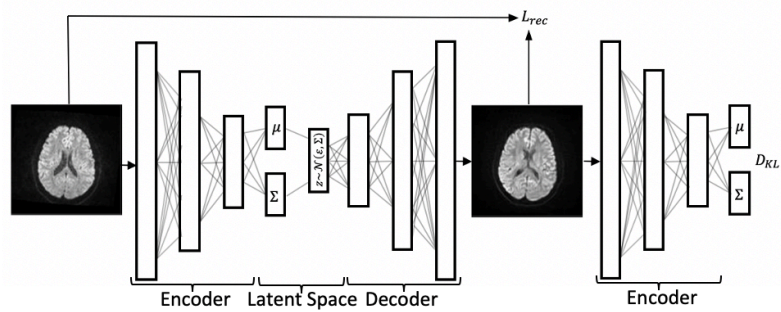
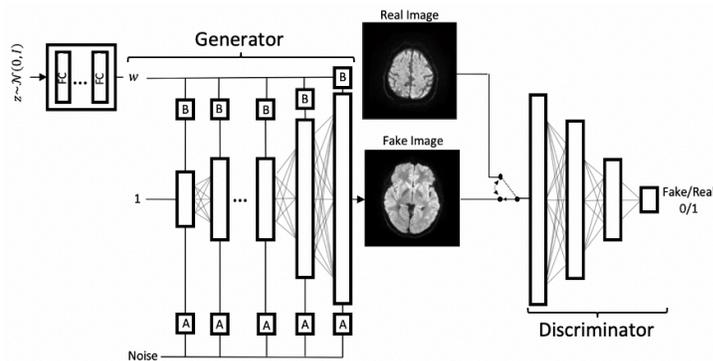

Figure 2. a) Deep Feature Consistent Variational Autoencoder. Original and reconstructed images are compared in the feature space by extracting hidden representations of both out of a pretrained VGG16. This procedure enforces spatial correlations to a higher extent than MSE in the pixel space. b) Introspective Variational Autoencoder. Encoder network is trained to reduce dimensionality of input images and act as a discriminator for synthetic MRI brains. Contrary to a typical GAN, the encoder does output a real/fake label, but evaluates this based on the similarity (in terms of Kullback-Leibler Divergence) between the probability distribution the encodings follow and a standard unit gaussian distribution. c) Style-Base Generative Adversarial Model. Input to the generator is no longer only gaussian noise, but a constant, a style vector resulting from transforming gaussian noise through a learned mapping function, and noise inserted at different layers.

## 3 Method

### 3.1 Database and Image Normalization

Institutional Review Board approval from Jeremy Heit and Eric Tranvinh was obtained. A database of 2,029 cases with normal brain (mean patient age $38 \pm 24$ y, 1,088 females, 939 males), as defined by a report with normal finding, were downloaded using an in-house PACS crawler. [25] Images were co-registered with an affine transformation to the standard MNI space rotated in the anterior commissure – posterior, and resampled to a standard resolution of 128 x 128 x 40 voxels, using ANTS [26]. The 6 top and 8 bottom slices were discarded, resulting in 52,754 brain images.

Signal intensity values were clipped at the 95th percentile and normalized by dividing by the remaining maximum pixel intensity value.

*3.2 Network Architectures and Training Details*

All four networks were trained on a NVIDIA GTX 1080 GPU Card.

*Vanilla Variational Autoencoder*

The network followed the structure as detailed in the Ref. [15]. The encoder consisted of 4 convolutional layers of $4 \times 4$ kernels. The stride was set to the constant value of 2 and batch normalization [27] was used before LeakyReLU activations. The encoder network concluded with two fully connected layers to output mean and variance of the encodings. The decoder was formed by 4 convolutional layers, this time with $3 \times 3$ kernels and a constant stride of 1. Upsampling was performed following the nearest neighbor method by a scale of 2. Batch normalization and a LeakyReLU activation again followed each convolution. The network was trained for 48 hours corresponding to 2000 epochs, with the Adam method as optimizer, a learning rate set to the constant value of 0.0001 and a batch size of 64. The value for the dimension of the latent vectors was set to 200 and the pixel-wise squared error term in equation (5) was scaled by $0.025 \times 128 \times 128$.

*Deep Feature Consistent Variational Autoencoder*

The implemented architecture for both encoder and decoder were identical to the one employed for the vanilla variational autoencoder. A pretrained VGG16, with the weights obtained from training on the ImageNet dataset and with exclusion of the top layers, was used to compute the feature perceptual loss: the outputs of three ReLU activation layers were extracted for this purpose. This network was trained for 48 hours, corresponding to 2000 epochs, with a batch size of 64, the Adam optimization method and a constant learning rate set to 0.0001. The dimension of the latent space was fixed to 200. The weighting of the feature perceptual loss was set to $2 \times 10^{-9}$.

*Introspective Variational Autoencoder*

The detailed guidelines presented in the original paper were followed to implement the architecture of the Introspective Variational Autoencoder. As explained there [16], encoder and generator were similar to the corresponding networks in the PGGAN [28] (Progressive Growing of GANs), but with the additional use of residual blocks [23], where each layer feeds into the next layer and directly into the layer 2-3 depths further. For our dataset image's dimension of 128x128, the hyperparameter *m* was set to 110 and the latent dimension to 256, $\mathcal{L}_{KL}(\cdot)$ terms in the loss functions (8,9) were scaled by the constant value of 0.25 and $\mathcal{L}_{rec}(\cdot,\cdot)$ terms by 0.5. Training was performed during 48 hours, corresponding to 300 epochs, with a batch size of 8, the Adam optimization method and a constant learning rate of 0.0002.

*Style-Based GAN*

We slightly modified the Style-Based GAN from the original paper [17] as follows: The mapping function was represented by 2 fully connected layers, with both input and output activation dimensions of 512. Progressive growing and mixing regularization were discarded for simplicity. Encoder and decoder were adopted from the implementation of Progressive GANs [28]. Both the discriminator and generator were trained for 48 hours, corresponding to a total of 200 epochs, optimized with the Adam method with a constant learning rate of 0.0001. All layer's weights were initialized with the He Normal Initializer.

### 3.3 Quantitative Evaluation Metrics

*Quantitative evaluation of the image reconstruction for the VAE networks*

In the case of the VAE networks, for which image reconstruction is possible, the quality of the image reconstruction was evaluated on a set of 1000 reconstructed images and compared to the original 1000 images, using the pixel-by-pixel-wise mean squared error. Additionally, the sharpness of the reconstructed image was compared to the original image using the Laplace Variance Score (defined below). GANs do not reconstruct images so this evaluation was only performed for the VAE networks.

*Quantitative evaluation of the image generation*

The following four metrics were evaluated on a set of 1000 generated images produced by the four networks and on the original dataset in order to have a reference value for comparison.

- *Dataset Similarity (DS)*: is the average pixel-wise squared error between the generated images and all the original images. A smaller value indicates larger similarity of the generated images to the original images, a larger value indicates a larger diversity of the generated images from the original images.

$$DS = \frac{1}{\#Samples} \sum_{X_{sample}} \sum_{X_{original}} \frac{1}{128 \times 128} \|X_{sample} - X_{original}\|_2^2 \quad (11)$$

- *Intra-Sample Diversity (ISD)*: is the average pixel-wise squared error between each generated image and all remaining generated images. The greater this quantity, the more diverse the generated images are with respect to each other.

$$ISD = \frac{1}{\#Samples} \sum_{X_{sample,i}} \sum_{\substack{X_{Sample,j} \\ i \neq j}} \frac{1}{128 \times 128} \|X_{sample,i} - X_{Sample,j}\|_2^2 \quad (12)$$

- *Minimum Intra-Sample Diversity (Minimum ISD)*: is the average pixel-wise squared error between each generated image and its nearest neighbor, where the nearest neighbor is defined as the sample closest to it in terms of the squared $L_2$ norm. This metric allows to quantify how similar a pair of most similar samples is on average.

$$Minimum\ ISD = \frac{1}{\#Samples} \sum_{X_{sample}} \frac{1}{128 \times 128} \|X_{sample} - n(X_{sample})\|_2^2,$$
$$with\ \ n(X) = \min_{j \in Samples\{X\}} \|X - X_j\|_2^2 \quad (13)$$

The smaller this value, the more similar two neighboring samples are, indicating higher redundancy in the pool of generated samples.

- *Laplace Variance Sharpness Score*: The sharpness of the generated images was evaluated using the Laplace Variance Score as follows: the Laplacian filter, a well-known edge detector, was applied to the image and the variance of the result was computed. A well-focused, sharp image has a high Laplacian Variance Sharpness Score, a blurry image a low Laplacian Variance Sharpness Score.

For the evaluation of all metrics, the images were used in their normalized uncropped form, i.e. the pixel intensity values were constrained to the interval [0,1].

*3.4 Subjective assessment*

A dataset of 250 images for the subjective evaluation was prepared as follows: 50 real images chosen randomly from the original dataset and 50 random images generated by each of the four networks studied here. Two experienced neuroradiologists (Jeremy Heit, 7 years of experience, and Eric Tranvinh, 7 years of experience) from external institutions, fully blinded to the whole study, classified each image of this dataset as real or fake. Inter-rater reliability was assessed using a Cohen's kappa coefficient [29].

**4 Results**

*Image reconstruction for the VAE networks*

All three VAE networks learned to reconstruct images with correct shapes and anatomical relationships, for example the shape of the brain, the midline symmetry, the position and shape of the ventricles, and gray-white matter differentiation can be observed (Figure 3). The images reconstructed by the vanilla VAE were relatively blurry, while the reconstructed images from the DFC and the IntroVAE were sharper. The pixel-by-pixel mean squared error of the reconstructed images was the lowest for the Vanilla VAE ($0.004 \pm 0.02$) followed by the IntroVAE ($0.008 \pm 0.03$) and the DFC VAE ($0.008 \pm 0.03$). The Laplace variance sharpness score increased from the Classical VAE ($349 \pm 160$), the DFC VAE ($715 \pm 254$) to the introVAE ($1,222 \pm 490$), compared with $1,495 \pm 659$ for the original dataset.

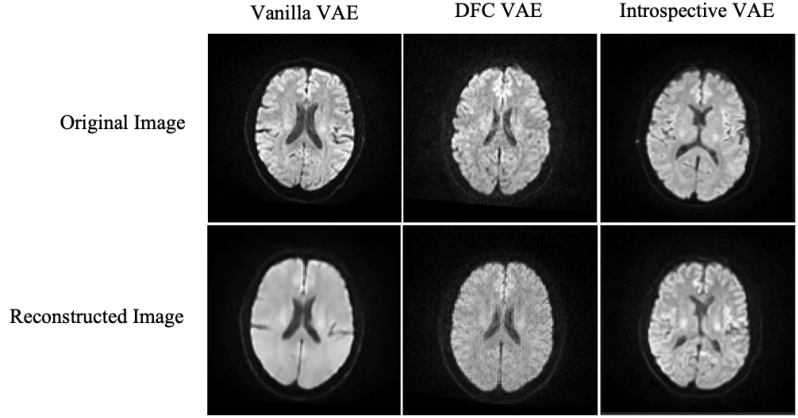

*Figure 3. Examples of original diffusion-weighted images and their corresponding reconstructed images by the decoders of the respective variational autoencoder networks. The image reconstructed by the vanilla VAE is relatively blurry, while the reconstructed images from the DFC and the Introspective VAE are sharper.*

*Image Generation*

All networks generated synthetic images with correct shapes and anatomical relationships. Images generated by the vanilla VAE and to a lesser extent by the DFC VAE were of relatively poor quality, while images generated by the introVAE and the modified style-based GAN were of much better general quality (Figure 4). As expected, the synthetic images generated by the vanilla VAE and the DFC VAE showed a low Laplace variance sharpness score (Table 1). The introVAE and the modified style-based GAN on the other side generated sharper images with a higher Laplace variance sharpness score as well as an ISD similar to the original images, but they generated sets of similar images with low Minimum ISD value (Table 1, Figure 4).

|  | Data Similarity | ISD | Minimum ISD | Laplace Sharpness Variance Score |
| --- | --- | --- | --- | --- |
| Original | 0.0494 | 0.0449 | 0.0063 | 1495 ± 659 |
| Vanilla VAE | 0.0552 | 0.0254 | 0.0078 | 201 ± 115 |
| DFC VAE | 0.0538 | 0.0219 | 0.0055 | 450 ± 219 |
| introVAE | 0.0478 | 0.0432 | 0.0009 | 1514 ± 560 |
| Style-Based GAN | 0.0487 | 0.0419 | 0.0008 | 1650 ± 715 |

*Table 1: Dataset Similarity, Intra-Sample Diversity, Minimum Intra-Sample Diversity and Laplace Variance Sharpness Score (mean ± standard deviation) evaluated on original images and samples generated by the four analyzed generative models. The Vanilla and DFC variational autoencoders generated samples that were quite similar to each other and of low quality. The introVAE and the modified style-based GAN were able to circumvent this problem, yet they generate sets of similar samples with low Minimum ISD value.*

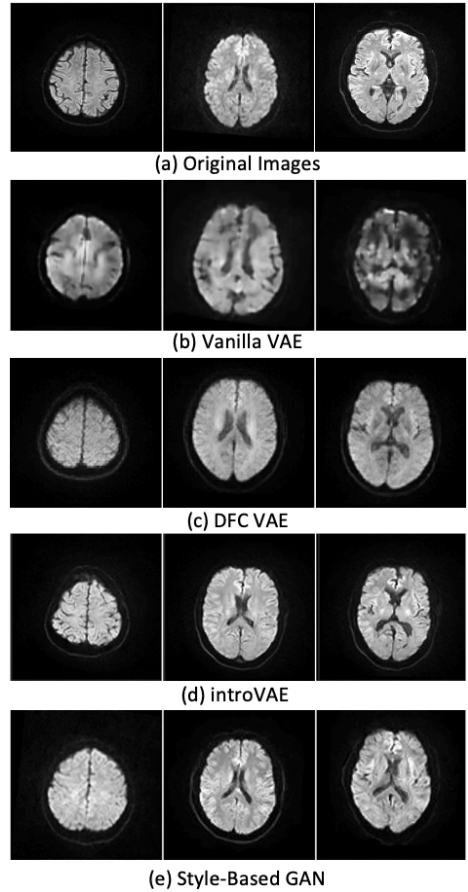

*Figure 4. Examples of (a) original and synthetic brain images generated by (b) Vanilla VAE, (c) DFC VAE, (d) introVAE and € style-based GAN, at three different levels in the brain. As can be seen in those examples, the generated images by the introVAE and the style-based GAN are of high quality.*

*Synthetic Image Generation: Subjective Evaluation*

The two expert neuroradiologists, blinded to the whole study, classified as expected the vast majority of the original images as real (([reader 1/ reader 2] 94% / 100%), as well as the vast majority of images generated by the introVAE and the style-based GAN like real original images (84% / 100% and 92% / 98%, respectively; Table 2), while the large majority of images generated by the Vanilla VAE and the DFC VAE were classified as fake (96% / 100% respectively 92% / 98%; Table 2). The inter-reader agreement was high (Cohen's kappa coefficient 0.84).

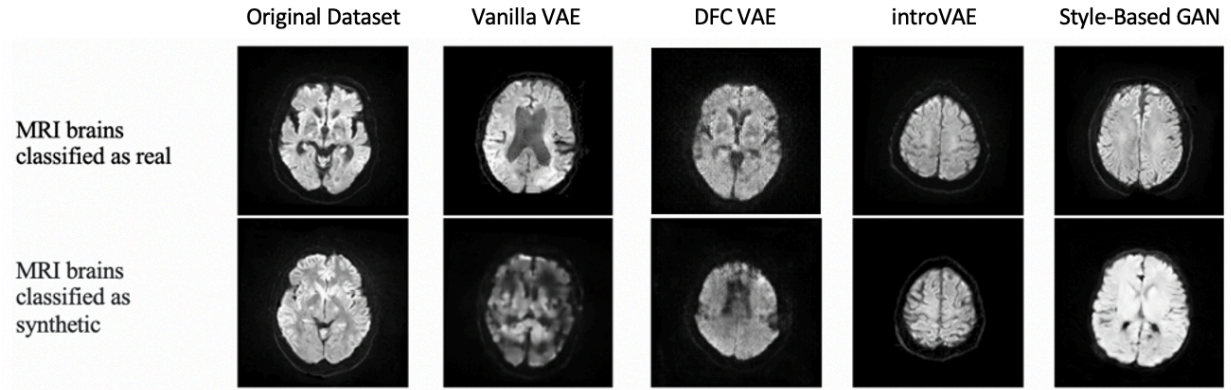

*Figure 5. Examples of correctly and incorrectly classified images by at least one of the experienced neuroradiologists. Upper row shows DWIs classified as real and lower row as fake.*

|  | Original Dataset | Vanilla VAE | DFC VAE | introVAE | Style-Base GAN |
|---|---|---|---|---|---|
| True Positives | 94% / 100% | 0% / 0% | 0% / 0% | 0% / 0% | 0% / 0% |
| False Positives | 0% / 0% | 4% / 0% | 0% / 4% | 84% / 100% | 92% / 98% |
| True Negatives | 0% / 0% | 96% / 100% | 100% / 96% | 16% / 0% | 8% / 2% |
| False Negatives | 6% / 0% | 0% / 0% | 0% / 0% | 0% / 0% | 0% / 0% |

*Table 2: Percentage of synthetic MR brain images generated by the four networks classified as real by the expert neuroradiologists. First entry corresponds to the first neuroradiologist, the second entry corresponds to the second expert. The introVAE and the modified style-based GAN generated images able to fool expert neuroradiologists to a high degree.*

## 5 Discussion

This study demonstrated that realistic synthetic diffusion-weighted images can be produced by an introVAE and a style-based GAN trained on a dataset of 50,000 images arising from 2,029 patients. The generated samples of these two models showed a data similarity close to the real image dataset, as well as a high data diversity, but the low Minimum ISD suggests that both networks generated groups of nearly identical images, suggesting mode collapse similar to what is observed for GAN-based models in natural images [14]. Both the vanilla VAE and the deep feature consistent VAE were capable of capturing topologic information such as overall shape and intensity, but the images generated by these two networks were easily classified as fake by the two expert neuroradiologists. The poorer image quality of the samples generated by the Vanilla VAE and the DFC VAE was quantitatively supported by a Laplace Variance Score of respectively around 7 and 3.5 times lower than for the real DWIs. An additional problem with these two networks is the low mean intra-sample diversity, suggesting the synthetic images created by these networks were more similar to each other than images taken from the original dataset.

While VAEs are designed to approximate the likelihood explicitly, GANs are designed to generate realistic images, thereby sampling from the true data probability distribution without knowing its explicit from [18]. As a consequence, and in accordance with the literature based on natural images, we found for diffusion-weighted images that standard VAE networks such as the vanilla VAE and the DFC VAE generate blurry ("averaged") images [9] while the GAN-based architectures achieved more realistic images of greater quality compared to VAEs [16]. While GANs produce sharper images, they are known to be more difficult to train [18, 30, 31], the reason being that the generator and the discriminator are trained simultaneously in an adversarial manner, meaning that improvements to one network come at the expense of the other. GANs are known to get stuck in modes even if the data is multi-modal, which refers to the fact that the generator only produces one or a small subset of different outcomes or modes [14]; a fact that is also observed in this study. Nonetheless, the high-quality MR images produced by the best performing underlying generative models show the ability of such networks to augment datasets of insufficient size with realistic samples. Kuzuhiro et al. [32], who investigated the level of realism of MR images synthesized with generative models, came to the same conclusion, and Sandfort et al. [4] further showed the positive impact of data augmented with this technique in improving generalizability in CT segmentation tasks. Bermudez at al. [33], focusing on one particular GAN, showed the capability of a specific GAN network to generate MRI images with high resemblance to those in the training set.

[34] further analyzed the use of generative models for data augmentation in the medical field, specifically applied to CTs of liver lesions.

There are several limitations to this study. One, this was a single center study and no data from other institutions or sources was used. Measuring quantitatively and objectively the quality and diversity of samples generated by generative models is a well-known challenge [18]. For natural images, the Inception Score [31] and the Fréchet Inception Distance [35] have been used. However, these metrics were developed for natural images, and their validity for medical images is under debate [36, 37]. Further, these metrics would require a labelled database, which was not available here, and it is not even clear what labels could be used in the cases of normal brain images, which don't have clear characteristic differences. Here we have introduced four indicative quantitative metrics to assess the sharpness, diversity between the samples, and dataset similarity for the evaluation of generated medical images, and we used blinded expert opinion, which most likely remains the best current metric [36] and has been the standard metric used in related work [33, 34].

In conclusion, we showed that realistic-looking diffusion-weighted magnetic resonance images can be synthesized using a Style-Based GAN or an introspective VAE, with comparable data similarity and sharpness similar to the original real dataset; however both yield a reduced minimum intra-sample diversity. Nonetheless, these findings promise to improve the applicability of deep learning in a medical context by significantly upscaling the available training data.